\documentclass[10pt,twocolumn,letterpaper]{article}

\usepackage{cvpr}
\usepackage{times}
\usepackage{epsfig}
\usepackage{graphicx}
\usepackage{amsmath}
\usepackage{amssymb}
\usepackage{url}
\usepackage{subcaption}
\usepackage[breaklinks=true,bookmarks=false]{hyperref}

\cvprfinalcopy 

\def\cvprPaperID{****} 

\ifcvprfinal\fi
\begin{document}

\title{Augmented Reality Meets Computer Vision : Efficient Data Generation for Urban Driving Scenes}

\author{Hassan Abu Alhaija$^{1}$ \quad Siva Karthik Mustikovela$^{1}$\\%
Lars Mescheder$^{2}$ \quad  Andreas Geiger$^{2,3}$ \quad Carsten Rother$^{1}$\\%
\\%
$^1$Computer Vision Lab, TU Dresden\\%
$^2$Autonomous Vision Group, MPI for Intelligent Systems T\"ubingen\\%
$^3$Computer Vision and Geometry Group, ETH Z\"urich\\%
}

\maketitle
\begin{abstract} 
The success of deep learning in computer vision is based on the availability
of large annotated datasets. To lower the need for hand labeled images, virtually rendered 3D worlds have recently gained popularity. Unfortunately, creating realistic 3D content is challenging on its own and requires significant human effort. In this work, we propose an alternative paradigm which combines real and synthetic data for learning semantic instance segmentation and object detection models. Exploiting the fact that not all aspects of the scene are equally important for this task, we propose to augment real-world imagery with virtual objects of the target category. Capturing real-world images at large scale is easy and cheap, and directly provides real background appearances without the need for creating complex 3D models of the environment. We present an efficient procedure to augment these images with virtual objects. This allows us to create realistic composite images which exhibit both realistic background appearance as well as a large number of complex object arrangements.
In contrast to modeling complete 3D environments, our data augmentation approach requires only a few user interactions in combination with 3D shapes of the target object category.
Through an extensive set of experiments, we conclude the right set of parameters to produce augmented data which can maximally enhance the performance of instance segmentation models.
Further, we demonstrate the utility of proposed approach on training standard deep models for semantic instance segmentation and object detection of cars in outdoor driving scenarios. We test the models trained on our augmented data on the KITTI 2015 dataset, which we have annotated with pixel-accurate ground truth, and on the Cityscapes dataset. Our experiments demonstrate that models trained on augmented imagery generalize better than those trained on synthetic data or models trained on limited amounts of annotated real data.
\end{abstract}
\vspace{-0.7cm}
\section{Introduction} 
\label{sec:introduction}

In recent years, deep learning has revolutionized the field of computer vision. Many tasks that seemed elusive in the past, can now be solved efficiently and with high accuracy using deep neural networks, sometimes even exceeding human performance (\cite{Taigman2014CVPR}).

However, it is well-known that training high capacity models such as deep neural networks requires huge amounts of labeled training data. This is particularly problematic for tasks where annotating even a single image requires significant human effort, \eg, for semantic or instance segmentation.
A common strategy to circumvent the need for human labels is to train neural networks on synthetic data obtained from a 3D renderer for which ground truth labels can be automatically obtained,  (\cite{Shafaei2016ARXIV,Richter2016ECCV,Movshovitz-Attias2016ECCVWORK,Varol2017ARXIV,Zhang2016ARXIVb,Ros2016CVPR,Handa2016CVPR,Gaidon2016CVPR}).
While photo-realistic rendering engines exist (\cite{Jakob2010}), the level of realism is often lacking fine details in the 3D world,  \eg, leaves of trees can only be modeled approximately.

\begin{figure*}[t!]
    \centering
	\includegraphics[width=\linewidth]{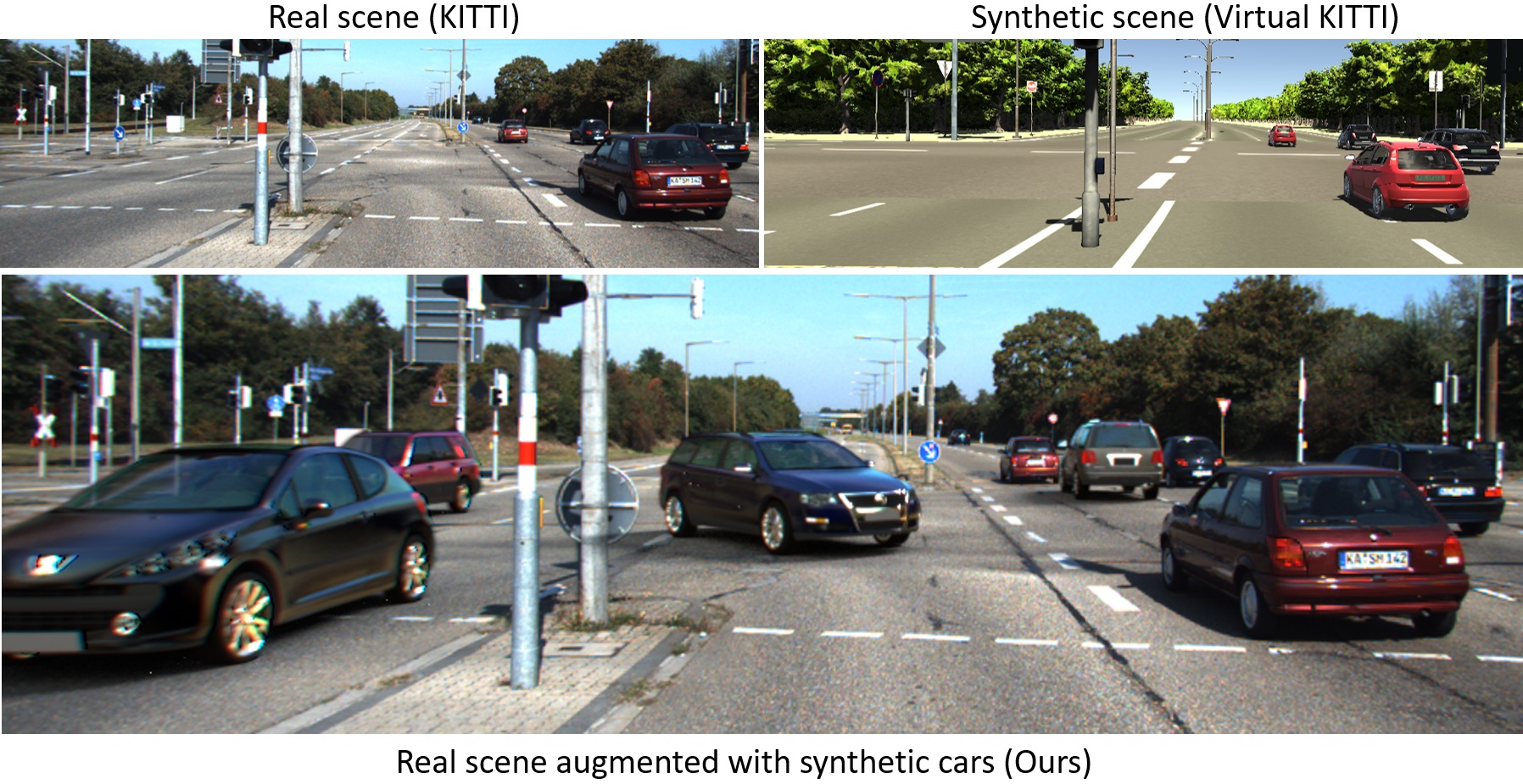}
    \caption{Obtaining  synthetic training data usually requires building large virtual worlds (top right) (\cite{Gaidon2016CVPR}). We propose a new way to extend datasets by augmenting real training images (top left) with realistically rendered cars (bottom) keeping the resulting images close to real while expanding the diversity of training data.
    	}
    \label{fig:teaser}
\end{figure*}

In this paper, we demonstrate that state-of-the-art photo-realistic rendering can be utilized to augment real-world images and obtain virtually unlimited amounts of training data for specific tasks such as semantic instance segmentation and object detection. Towards this goal, we consider real images with additional side information, such as camera calibration and environment maps, and augment these images with novel object instances. In particular, we augment the data with realistically rendered car instances.  This allows us to keep the full realism of the background while being able to generate arbitrary amounts of foreground object configurations.

Figure \ref{fig:teaser} shows a real image before and after augmentation.
While our rendered objects rival the realism of the input data, they provide the variations (\eg, pose, shape, appearance) needed for training deep neural networks for  instance aware semantic segmentation and bounding box detection of cars.
By doing so, we are able to considerably improve the accuracy of state-of-the-art deep neural networks trained on real data.

While the level of realism is an important factor when synthesizing new data, there are two other important aspects to consider - data diversity and human labor. Manually assigning a class or instance label to every pixel in an image is possible but tedious, requiring up to one hour per image (\cite{Cordts2016CVPR}).
Thus existing real-world datasets are limited to a few hundred (\cite{Brostow2009PRL}) or thousand (\cite{Cordts2016CVPR}) annotated examples, thereby severely limiting the diversity of the data. In contrast, the creation of virtual 3D environments allows for arbitrary variations of the data and virtually infinite number of training samples.
However, the creation of 3D content requires professional artists and the most realistic 3D models (designed for modern computer games or movies) are not publicly available due to the enormous effort involved in creating them.
While \cite{Richter2016ECCV} have recently demonstrated how content from commercial games can be accessed through manipulating low-level GPU instructions, legal problems are likely to arise and often the full flexibility of the data generation process is no longer given. 

In this work we demonstrate that the creation of an augmented dataset which combines real with synthetic data requires only moderate human effort while yielding the variety of data necessary for improving the accuracy of
state-of-the-art instance segmentation network (Multitask Network Cascades) (\cite{Dai2016CVPR}) and object detection network (Faster R-CNN) (\cite{ren2015faster}).
In particular, we show that a model trained using our augmented dataset generalizes better than models trained purely on synthetic data as well as models which use a smaller number of manually annotated real images.
Since our data augmentation approach requires only minimal manual effort,
we believe that it constitutes an important milestone towards the ultimate task of creating virtually infinite, diverse and realistic datasets with ground truth annotations. 
In summary, our contributions are as follows:
\begin{itemize}
\item We propose an efficient solution for augmenting real images with photo-realistic synthetic object instances which can be arranged in a flexible manner.
\item We provide an in-depth analysis of the importance of various factors of the data augmentation process, including the number of augmentations per real image, the realism of the background and the foreground regions.
\item We find that models trained on augmented data generalize better than models trained on purely synthetic data or small amounts of labeled real data.
\item For conducting the experiments in this paper, we introduce two newly labeled instance segmentation datasets, named KITTI-15 and KITTI-360, with a total of 400 images.
\end{itemize}

\section{Related Work} 
\label{sec:related}

Due to the scarcity of real-world data for training deep neural networks, several researchers have proposed to use synthetic data created with the help of a 3D rendering engine. 
Indeed, it was shown (\cite{Shafaei2016ARXIV,Richter2016ECCV,Movshovitz-Attias2016ECCVWORK}) that deep neural networks can achieve state-of-the-art results when trained on synthetic data and that the accuracy can be further improved by fine tuning on real data (\cite{Richter2016ECCV}).
Moreover, it was shown that the realism of synthetic data is important to obtain good performance (\cite{Movshovitz-Attias2016ECCVWORK}). 

Making use of this observation, several synthetic datasets have been released which we will briefly review in the following.
\cite{Hattori2015CVPR} present a scene-specific pedestrian detector using only synthetic data. \cite{Varol2017ARXIV} present a synthetic dataset of human bodies and use it for human depth estimation and part segmentation from RGB-images. In a similar effort, \cite{Chen2016THREEDV} use synthetic data for 3D human pose estimation. In (\cite{Souza2016ARXIV}), synthetic videos are used for human action recognition with deep networks.
\cite{Zhang2016ARXIVa} present a synthetic dataset for indoor scene understanding. Similarly, \cite{Handa2016CVPR} use synthetic data to train a depth-based pixelwise semantic segmentation method. In \cite{Zhang2016ARXIVb}, a synthetic dataset for stereo vision is presented which has been obtained from the UNREAL rendering engine. \cite{Zhu2016ARXIV} present the AI2-THOR framework, a 3D environment and physics engine which they leverage to train an actor-critic model using deep reinforcement learning.
\cite{Peng2015ICCV} investigate how missing low-level cues in 3D CAD models affect the performance of deep CNNs trained on such models. \cite{Stark2010BMVC} use 3D CAD models for learning a multi-view object class detector.

In the context of autonomous driving, the SYNTHIA dataset (\cite{Ros2016CVPR}) contains a collection of diverse urban scenes and dense class annotations. \cite{Gaidon2016CVPR} introduce a synthetic video dataset (Virtual KITTI) which was obtained from the KITTI-dataset (\cite{Geiger2013IJRR}) alongside with dense class annotations, optical flow and depth. \cite{Su2015ICCVa} use a dataset of rendered 3D models on random real images for training a CNN on viewpoint estimation. While all aforementioned methods require labor intensive 3D models of the environment, we focus on exploiting the synergies of real and synthetic data using augmented reality. In contrast to purely synthetic datasets, we obtain a large variety of realistic data in an efficient manner. Furthermore, as evidenced by our experiments, combining real and synthetic data within the same image results in models with better generalization performance.

While most works use either real or synthetic data, only few papers consider the problem of training deep models with mixed reality. \cite{Rozantsev2015CVIU} estimate the parameters of a rendering pipeline from a small set of real images for training an object detector.
\cite{Gupta2016CVPR} use synthetic data for text detection in images.
\cite{Pishchulin2011CVPR} use synthetic human bodies rendered on random backgrounds for training a pedestrian detector. \cite{Dosovitskiy2015ICCV} render flying chairs on top of random Flickr backgrounds to train a deep neural network for optical flow.
Unlike existing mixed-reality approaches which are either simplistic, consider single objects or augment objects in front of random backgrounds, our goal is to create high fidelity augmentations of complex multi-object scenes at high resolution.
In particular, our approach takes the geometric layout of the scene, environment maps as well as artifacts stemming from the image capturing device into account. We experimentally evaluate which of these factors are important for training good models.

\section{Data Augmentation Pipeline}
\label{sec:method}

\begin{figure*}[t!]
\centering
\includegraphics[width=\textwidth]{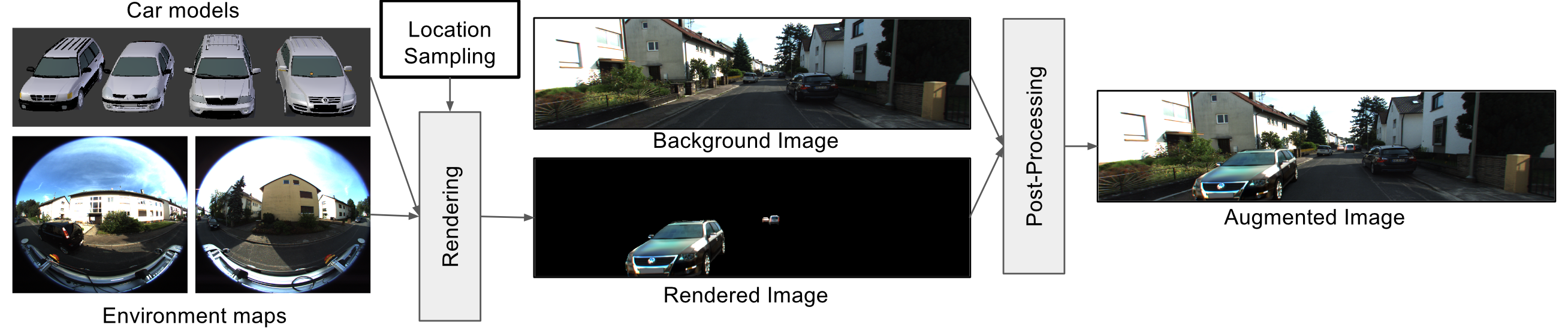}
\caption{Overview of our augmentation pipeline. Given a set of 3D car models, locations and environment maps, we render high quality cars and overlay them on top of real images. The final post-processing step insures better visual matching between the rendered and real parts of the resulting image.}
\label{fig:method} 
\end{figure*}

In this section, we describe our approach to data augmentation through photo-realistic rendering of 3D models on top of real scenes. 
To achieve this, three essential components are required: (i) detailed high quality 3D models of cars, (ii) a set of 3D locations and poses used to place the car models in the scene and, (iii) the environment map of the scene that can be used to produce realistic reflections and lighting on the models that matches the scene.

\begin{figure}[h!]
	\centering
	\begin{subfigure}[t]{\columnwidth}
	\centering
	\includegraphics[width=\columnwidth]{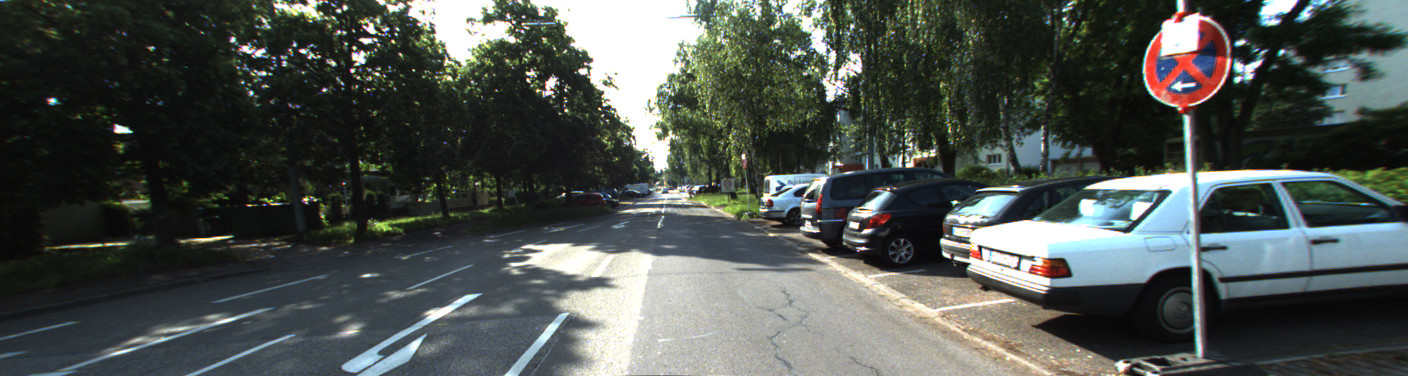}
	\end{subfigure}%
	\\
	\begin{subfigure}[t]{\columnwidth}
	\centering
	\includegraphics[width=\columnwidth]{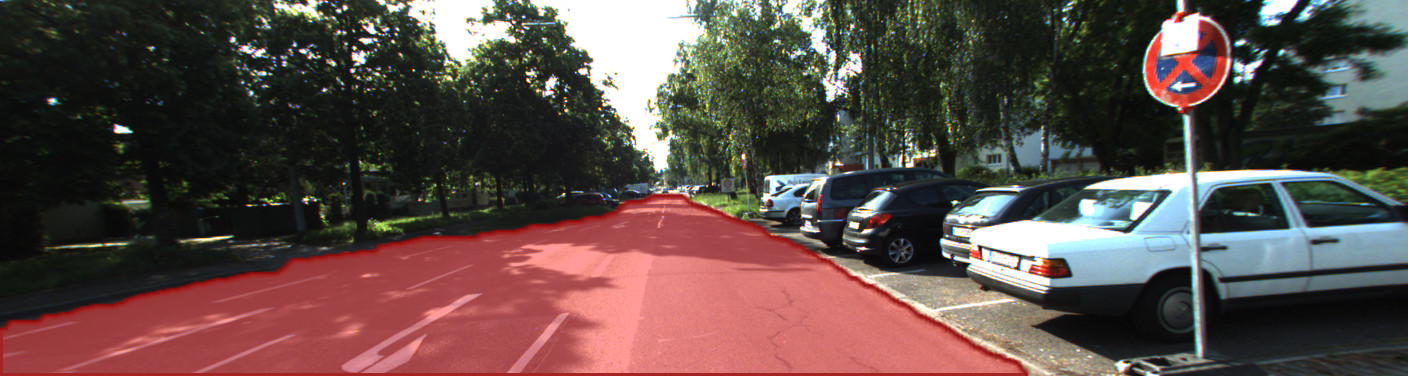}
	\end{subfigure}%
	\\
	\begin{subfigure}[t]{0.8\columnwidth}
	\centering
	\includegraphics[width=0.8\columnwidth]{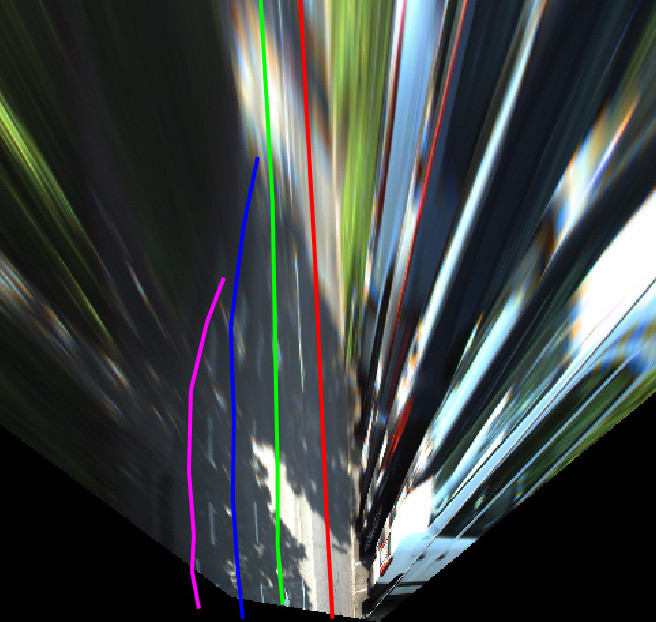}
	\end{subfigure}%
	
	\caption{(Top) The original image. (Middle) Road segmentation using (\cite{Teichmann2016ARXIV}) in red for placing synthetic cars. (Down) Using the camera calibration, we project the ground plane to get a birdseye view of the scene. From this view, the annotator draws lines indicating vacant trajectories where synthetic cars can be placed.}
	\label{fig:birdseye} 
\end{figure}

We use 28 high quality 3D car models covering 7 categories (SUV, sedan, hatchback, station wagon, mini-van, van) obtained from online model repositories\footnote{\url{http://www.dmi-3d.net}}. The car color is chosen randomly during rendering to increase the variety in the data.
To achieve high quality realistic augmentation, it is essential to correctly place virtual objects in the scene at practically plausible locations, matching the distribution of poses and occlusions in the real data. We explored four different location sampling strategies: (i) Manual car location annotations, (ii) Automatic road segmentation, (iii) Road plane estimation, (iv) Random unconstrained location sampling. 
For (i), we leverage the homography between the ground plane and the image plane, transforming the perspective image into a birdseye view of the scene. Based on this new view, our in-house annotators marked possible car trajectories (Figure \ref{fig:birdseye}). We sample the locations from these annotations and set the rotation along the vertical axis of the car to be aligned with the trajectory. 
For (ii), we use the algorithm proposed by (\cite{Teichmann2016ARXIV}) which segments the image into road and non-road areas with high accuracy. We back-project those road pixels and compute their location on the ground plane to obtain possible car locations and use a random rotation around the vertical axis of the vehicle.  While this strategy is simpler, it can lead to visually less realistic augmentations.
For (iii), since we know the intrinsic parameters of the capturing camera and its exact pose, it is possible to estimate the ground plane in the scene. This reduces the problem of sampling the pose from 6D to 3D, namely the 2D position on the ground plane and one rotation angle around the model's vertical axis.
Finally for (iv), we randomly sample locations and rotations from an arbitrary distribution. 

We empirically found Manual car location annotations to perform slightly better than Automatic road segmentation and on par with road plane estimation as described in Sec.~\ref{sec:evaluation}. We use manual labeling in all our experiments, unless stated otherwise.

\begin{figure*}[t]
    \centering
    \begin{subfigure}[t]{0.9\textwidth}
        \centering
        \includegraphics[width=\textwidth]{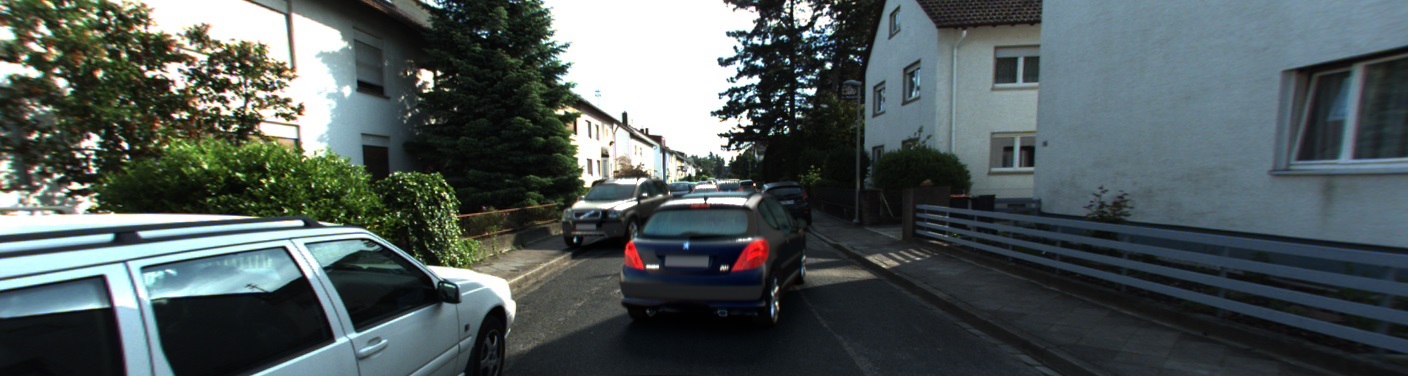}
        \caption{The two cars in the center are rendered}
    \end{subfigure}%
    \\
    \begin{subfigure}[t]{0.9\textwidth}
        \centering
        \includegraphics[width=\textwidth]{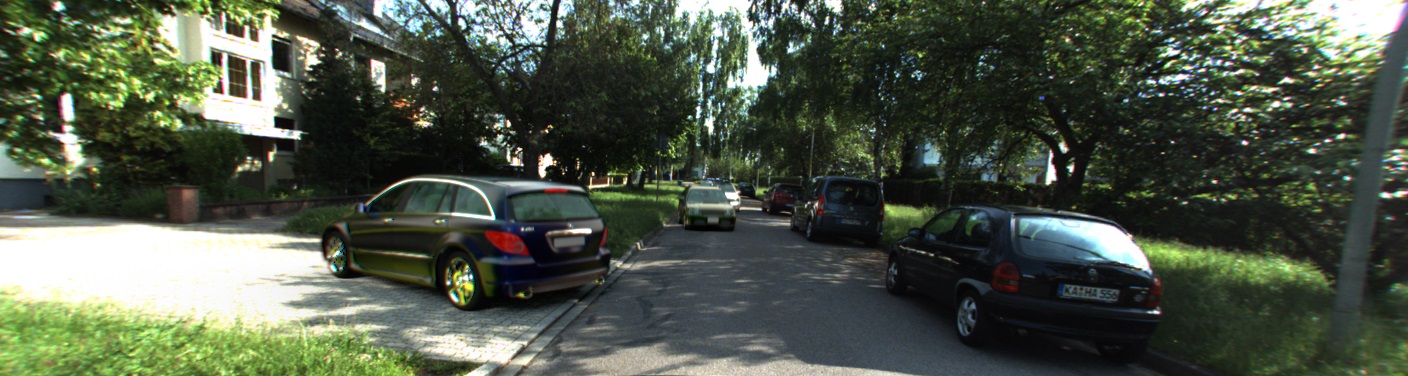}
        \caption{The car to the left and in the center are rendered}
    \end{subfigure}
    \\
    \begin{subfigure}[t]{0.9\textwidth}
        \centering
        \includegraphics[width=\textwidth]{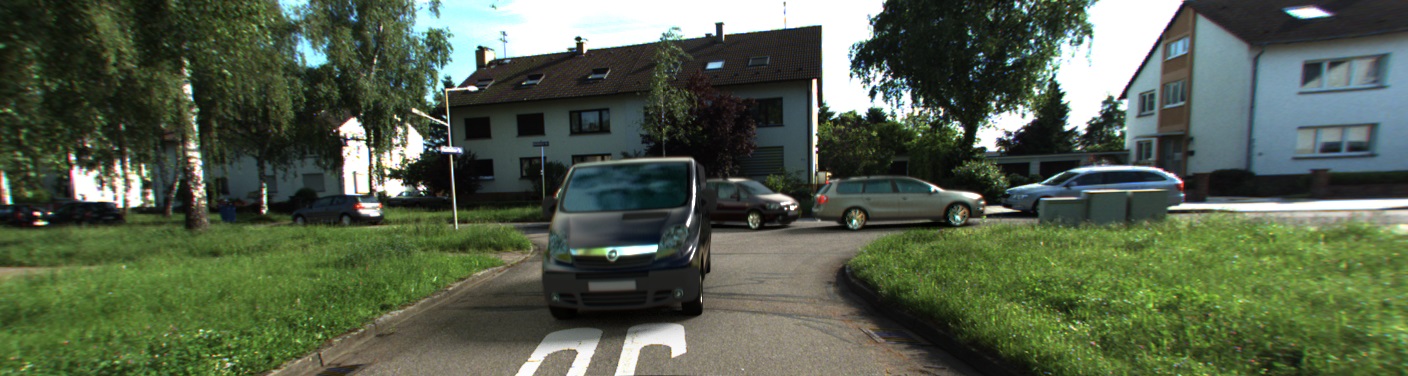}
        \caption{The three cars in the center are rendered}
    \end{subfigure}%
    \\
    \begin{subfigure}[t]{0.9\textwidth}
        \centering
        \includegraphics[width=\textwidth]{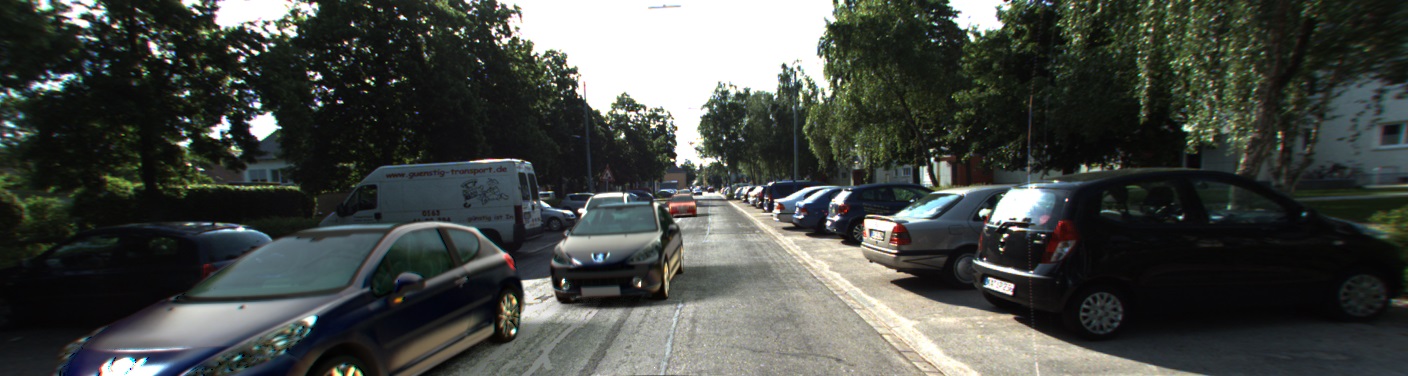}
        \caption{The three cars on the road are rendered}
    \end{subfigure}    
    \caption{Example images produced by our augmentation pipeline.}
    \label{fig:render_samples} 
\end{figure*}

We leverage the 360 degree panoramas of the environment from the KITTI-360 dataset (\cite{Xie2016CVPR}) as an environment map proxies for realistic rendering of cars in street scenes.
Using the 3D models, locations and environment maps, we render cars using the Cycle renderer implemented in Blender (\cite{blender}). Figure \ref{fig:method} illustrates our rendering approach. 
However, the renderings obtained from Blender lack typical artifacts of the image formation process such as motion blur, lens blur, chromatic aberrations, etc. To better match the image statistics of the background, we thus design a post-processing work-flow in Blender's compositing editor which applies a sequence of 2D effects and transformations to simulate those effects, resulting in renderings that are more visually similar to the background. More specifically, we apply color shifts to simulate chromatic aberrations in the camera lens as well as depth-blur to match the camera depth-of-field. Finally, we use several color curve and Gamma transformations to better match the color statistics and contrast of the real data. The parameters of these operations have been estimated empirically and some results are shown in Figure \ref{fig:render_samples}.

\section{Evaluation} 
\label{sec:evaluation}

In this section we show how augmenting driving scenes with synthetic cars is an effective way to expand a dataset and increase its quality and variance. In particular, we highlight two aspects in which data augmentation can improve the real data performance. First, introducing new synthetic cars in each image with detailed ground truth labeling makes the model less likely to over-fit to the small amount of real training data and exposes it to a large variety of car poses, colors and models that might not exist or be rare in real images. Second, our augmented cars introduce realistic occlusions of real cars which makes the learned model more robust to occlusions since it is trained to detect the same real car each time with a different occlusion configuration. This second aspect also protects the model from over-fitting to the relatively small amount of annotated real car instances. 

We study the performance of our data augmentation method on two challenging vision tasks, instance segmentation and object detection. Using different setups of our augmentation method, we investigate how the quality and quantity of augmented data affects the performance of a state-of-the-art instance segmentation model. In particular, we explore how the number of augmentations per real image and number of added synthetic cars affects the quality of the learned models. We compare our results on both tasks to training on real and fully synthetic data, as well as a combination of the two (i.e., training on synthetic data and fine-tuning on real data). We also experiment with different aspects of realism such as environment maps, post-processing and car placement methods.

\subsection{Datasets}
\paragraph{KITTI-360} For our experiments, we created a new dataset which contains 200 images chosen from the dataset presented in \cite{Xie2016CVPR}. We labeled all car instances at pixel level using our in-house annotators to create high quality semantic instance segmentation ground truth. This new dataset (KITTI-360) is unique compared to KITTI (\cite{Geiger2013IJRR}) or Cityscapes (\cite{Cordts2016CVPR}) in that each frame comes with two $180^{\circ}$ images taken by two fish-eye cameras on top of recording platform. Using an equirectangular projection, the two images are warped and combined to create a full $360^{\circ}$ omni-directional image that we use as an environment map during the rendering process. These environment maps are key to creating photo-realistic augmented images and are used frequently in Virtual Reality and Cinematic special effects applications. 
The dataset consists of 200 real images which form the basis for augmentation in all our experiments, i.e., we reuse each image $n$ times with differently rendered car configurations to obtain an $n$-fold augmented dataset.
\paragraph{VKITTI} To compare our augmented images to fully synthetic data, we use the Virtual KITTI (VKITTI) dataset (\cite{Gaidon2016CVPR}) which has been designed as a virtual proxy for the KITTI 2015 dataset (\cite{Menze2015CVPR}). Thus, the statistics of VKITTI (\eg, semantic class distribution, car poses and environment types) closely resembles those of KITTI-15 which we use as a testbed for evaluation. The dataset comprises $\sim$12,000 images divided into 5 sequences with 6 different weather and lighting conditions for each sequence.
\paragraph{KITTI-15} To demonstrate the advantage of data augmentation for training robust models, we create a new benchmark test dataset different from the training set using the popular KITTI 2015 dataset (\cite{Menze2015CVPR}). More specifically, we annotated all the 200 publicly available images of the KITTI 2015 (\cite{Menze2015CVPR}) with pixel-accurate semantic instance labels using our in-house annotators. While the statistics of the KITTI-15 dataset are similar to those of the KITTI-360 dataset, it has been recorded in a different year and at a different location / suburb. This allows us to assess performance of instance segmentation and detection methods trained on the KITTI-360 and VKITTI dataset.
\paragraph{Cityscapes} To further evaluate the generalization performance of augmented data, we test our models using the larger Cityscapes validation dataset (\cite{Cordts2016CVPR}) which consists of 500 instance mask annotated images. The capturing setup and data statistics of this dataset is different to those of KITTI-360, KITTI-15 and VKITTI making it a more challenging test set.

\subsection{Evaluation Protocol}

We evaluate the effectiveness of augmented data for training deep neural networks using two challenging tasks, instance-level segmentation and bounding-box object detection. In particular, we focus on the task of car instance segmentation and detection as those dominate our driving scenes.

\paragraph{Instance segmentation}
We choose the state-of-the-art Multi-task Network Cascade (MNC) by \cite{Dai2016CVPR} for instance-aware semantic segmentation. We initialize each model using the features from the VGG model (\cite{Simonyan2015ICLR}) trained on ImageNet and train the method using variants of real, augmented or synthetic training data. For each variant, we train the model until convergence and average the result of the best performing 5 snapshots on each test set. We report the standard average precision metric of an intersection-over-union threshold of $50\%$ (AP50) and $70\%$ (AP70), respectively. 
\paragraph{Object detection}
For bounding-box car detection we adopt the popular Faster-RCNN (\cite{ren2015faster}) method. We initialize the model using the VGG model trained on ImageNet as well and then train it using the same dataset variants for 10 epochs and average the best performing 3 snapshots on each test set. For this task, we report the mean average precision (mAP) metric commonly used in object detection evaluation. 

\begin{figure*}[t]
	\centering   
	\begin{subfigure}[b]{0.49\textwidth}
		\includegraphics[width=\textwidth]{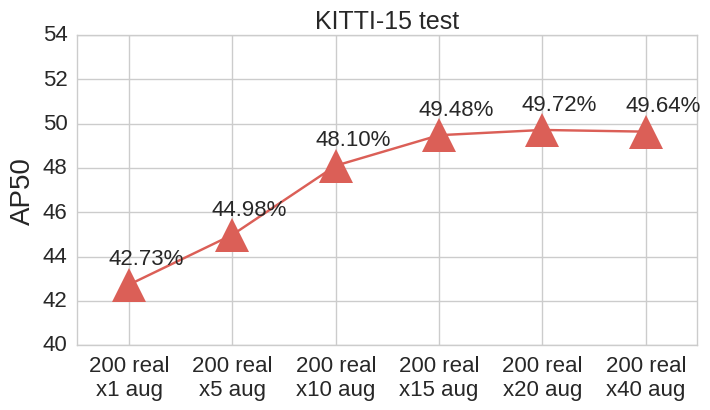}
		\caption{ }
		\label{fig:num_aug}
	\end{subfigure}
	\begin{subfigure}[b]{0.49\textwidth}
		\centering
		\includegraphics[width=\textwidth]{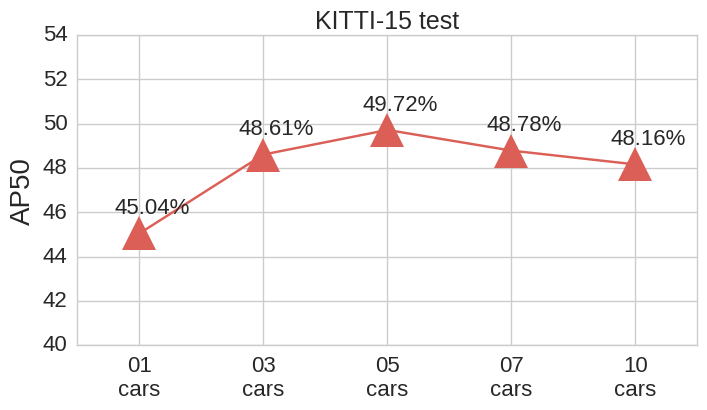}
		\caption{ }
		\label{fig:num_cars}
	\end{subfigure}
	\caption{Instance segmentation performance using augmented data. 
	(a) We fix the number of synthetic cars to 5 per augmentation and vary the number of augmentations per real image.
	(b) We fix the number of augmentations to 20 and vary the maximum number of synthetic cars rendered in each augmented image.}
	\label{fig:aug_analysis} 
\end{figure*}

\subsection{Augmentation Analysis}

We experiment with the two major factors for adding variation in the augmented data. Those are, (i) the number of augmentations, i.e the number of augmented images created from each real image, (ii) the number of synthetic cars rendered in each augmented images.\\
Figure \ref{fig:num_aug} shows how increasing the number of augmentations per real image improves the performance of the trained model through the added diversity of the target class, but then saturates beyond 20 augmentations.
While creating one augmentation of the real dataset adds a few more synthetic instances to each real image, it fails to improve the model performance compared to training on real data only. Nevertheless, creating more augmentations results in a larger and more diverse dataset that performs significantly better on the real test data. This suggests that the main advantage of our data augmentation comes from adding realistic diversity to existing datasets through having several augmented versions of each real image. In the rest of our experiments, we use 20 augmentations per real unless stated otherwise. 

In figure \ref{fig:num_cars} we examine the role of the synthetic content of each augmented image on performance by augmenting the dataset with various numbers of synthetic cars in each augmented image. At first, adding more synthetic cars improves the performance by introducing more instances to the training set. It provides more novel car poses and realistic occlusions on top of real cars leading to more generalizable models. Nevertheless, increasing the number of cars beyond 5 per image results in a noticeable decrease in performance. Considering that our augmentation pipeline works by overlaying rendered cars on top of real images, adding a larger number of synthetic cars will cover more of the smaller real cars in the image reducing the ratio of real to synthetic instances in the dataset. This negative effect soon undercuts the benefit of the diversity provided by the augmentation leading to decreasing performance.  
Our conjecture is that the best performance can be achieved using a balanced combination of real and synthetic data.
Unless explicitly mentioned otherwise, all our experiments were conducting using 5 synthetic cars per augmented image. 

\subsection{Comparing Real, Synthetic and Augmented Data}

\begin{figure*}[t]
	\centering
	\begin{subfigure}[b]{0.48\textwidth}
		\centering
		\includegraphics[width=\textwidth]{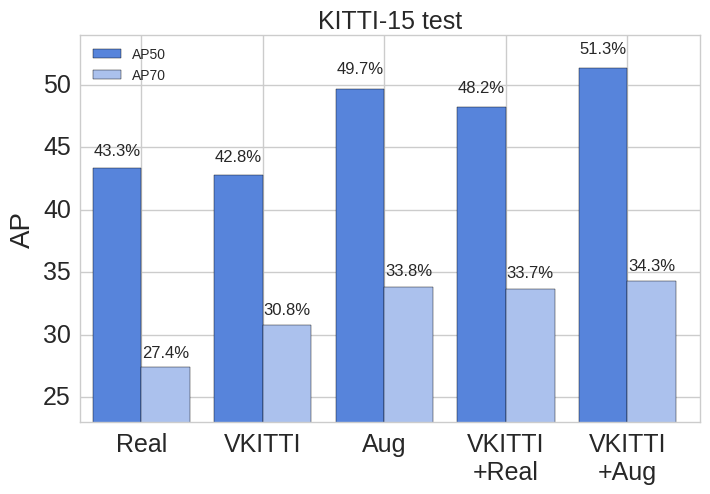}
		\caption{ }
		\label{fig:aug_vs_vkitti}
	\end{subfigure}
	\begin{subfigure}[b]{0.48\textwidth}
		\centering
		\includegraphics[width=\textwidth]{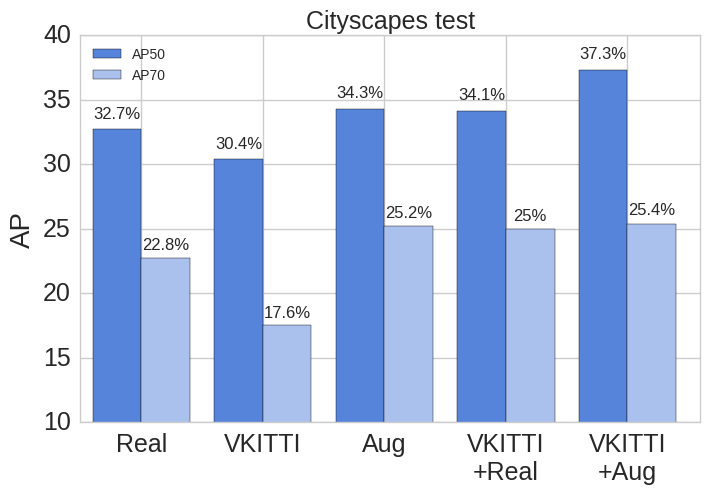}
		\caption{ }
		\label{fig:aug_vs_vkitti_cs}
	\end{subfigure}
	\caption{ Using our augmented dataset, we can achieve  better performance on both (a) the KITTI-15 test set and (b) Cityscapes (\cite{Cordts2016CVPR}) test set compared to using synthetic data or real data separately. We also outperform models trained on synthetic data and fine-tuned with real data (VKITTI+Real) while significantly reducing manual effort. Additionally, fine-tuning the model trained on VKITTI using our Augmented data (VKITTI+Aug) further improves the performance.
	}
	\label{fig:mnc_results} 
\end{figure*}
\begin{figure*}[t]
	\centering
	\begin{subfigure}[b]{0.48\textwidth}
		\centering
		\includegraphics[width=\textwidth]{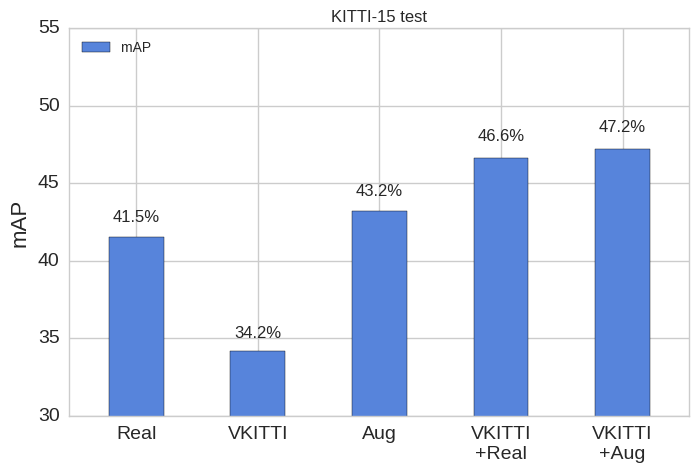}
		\caption{ }
		\label{fig:rcnn_aug_vs_vkitti}
	\end{subfigure}
	\begin{subfigure}[b]{0.48\textwidth}
		\centering
		\includegraphics[width=\textwidth]{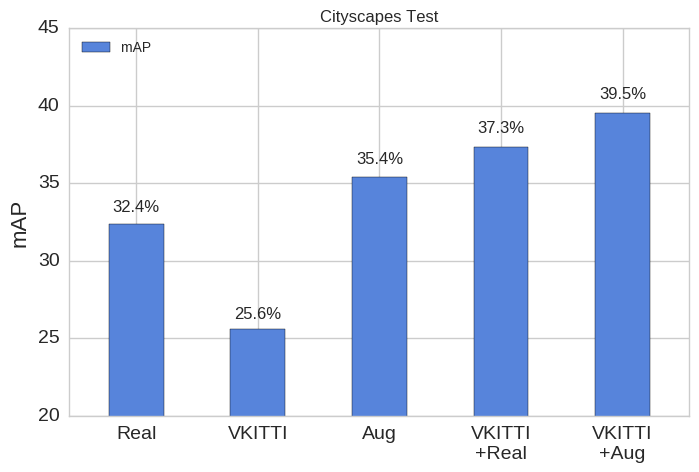}
		\caption{ }
		\label{fig:rcnn_aug_vs_vkitti_cs}
	\end{subfigure}
	
	\caption{Training the Faster RCNN model (\cite{ren2015faster}) for bounding box detection on various datasets. Using our augmented dataset we outperform the models trained using synthetic data or real data separately on both (a) KITTI-15 test set and (b) Citescapes (\cite{Cordts2016CVPR}) test set. We also outperform the model trained on VKITTI and fine-tuned on real data (VKITTI+Real) by using our augmented data to fine tune the model trained on VKITTI (VKITTI+Aug).
	}
	\label{fig:rcnn_results} 
\end{figure*}

Synthetic data generation for autonomous driving has shown promising results in the recent years. However, it comes with several drawbacks:
\begin{itemize}
\item The time and effort needed to create a realistic and detailed 3D world and populate it with agents that can move and interact. 
\item The difference in data distribution and pixel-value statistics between the real and virtual data prevents it from being a direct replacement to real training data. Instead, it is often used in combination with a two stage training procedure where the model is first pre-trained on large amounts of virtual data and then fine tuned on real data to better match the test data distribution. 
\end{itemize}
Using our data augmentation method we hope to overcome these two limitations. First, by using real images as background, we limit the manual effort to modeling high quality 3D cars compared to designing full 3D scenes. A large variety of 3D cars is available through online 3D model warehouses and can be easily customized. Second, by limiting the modification of the images to the foreground objects and compositing them with the real backgrounds, we keep the difference in appearance and image artifacts at minimum. As a result, we are able to boost the performance of the model directly trained on the augmented data without the need for a two stage pre-training/refinement procedure.

To compare our augmented data to fully synthetic data, we train a model using VKITTI and refine it with the real KITTI-360 training set. Figures \ref{fig:aug_vs_vkitti} and \ref{fig:aug_vs_vkitti_cs} show our results tested on KITTI-15 and Cityscapes respectively.
While fine-tuning a model trained on VKITTI with real data improves the results from 42.8\% to 48.2\%, our augmented dataset achieves a performance of 49.7\% in a single step. 
Additionally, using our augmented data for fine-tuning the VKITTI trained model significantly improves the results (51.3\%).
This demonstrates that the augmented data is closer in nature to real than to synthetic data. While the flexibility of synthetic data can provide important variability, it fails to provide the expected boost over real data due to differences in appearance. On the other hand, augmented data complements this by providing high visual similarity to the real data, yet preventing over-fitting.

While virtual data captures the semantics of the real world, at the low level real and synthetic data statistics can differ significantly. Thus training with purely synthetic data leads to biased models that under-perform on real data.
Similarly training or fine-tuning on a limited size dataset of real images restricts the generalization performance of the model. In contrast, the composition of real images and synthetic cars into a single frame can help the model to learn shared features between the two data distributions without over-fitting to the synthetic ones. 
Note that our augmented dataset alone performs slightly better than the models trained on VKITTI and fine-tuned on the real dataset only. This demonstrates that state-of-the-art performance can be obtained without designing complete 3D models of the environment.
Figure \ref{fig:rcnn_aug_vs_vkitti} and \ref{fig:rcnn_aug_vs_vkitti_cs} show similar results achieved for the detection task on both KITTI-15 and Cityscapes respectively. 

\subsection{Dataset Size And Variability}

\begin{figure*}[t]
	\centering   
	\begin{subfigure}[b]{0.48\textwidth}
		\includegraphics[width=\textwidth]{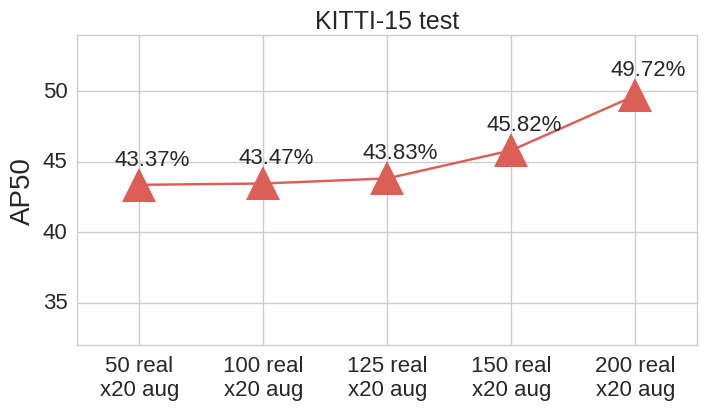}
		\caption{ }
		\label{fig:num_real_aug}
	\end{subfigure}
	\begin{subfigure}[b]{0.48\textwidth}
		\centering
		\includegraphics[width=\textwidth]{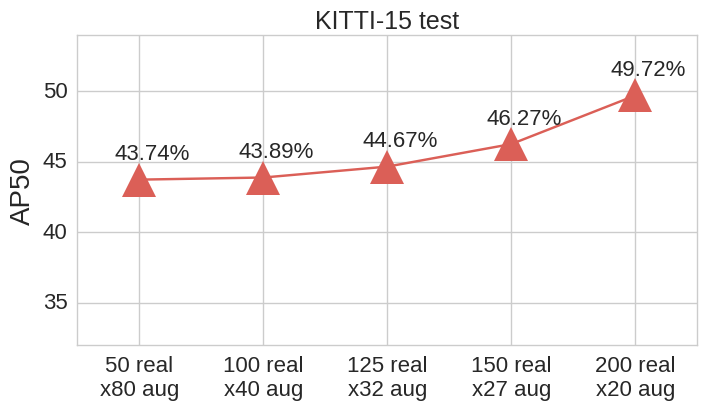}
		\caption{ }
		\label{fig:num_bg}
	\end{subfigure}
	\begin{subfigure}[b]{0.48\textwidth}
		\centering
		\includegraphics[width=\textwidth]{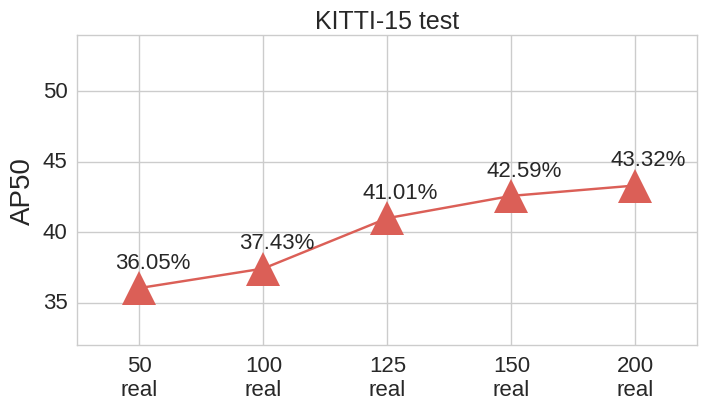}
		\caption{ }
		\label{fig:num_real}
	\end{subfigure}
	\begin{subfigure}[b]{0.48\textwidth}
		\centering
		\includegraphics[width=\textwidth]{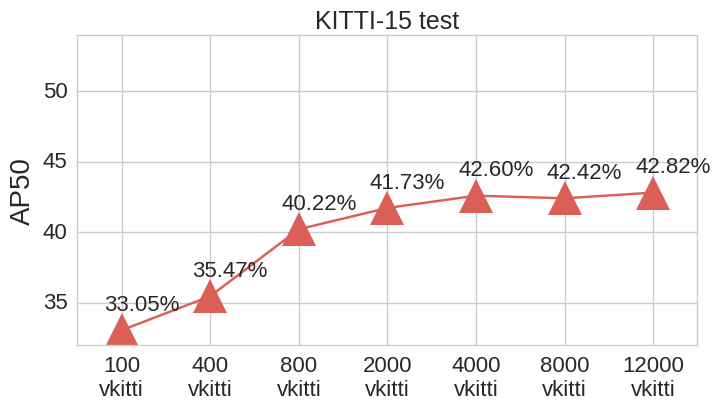}
		\caption{ }
		\label{fig:num_vkitti}
	\end{subfigure}
	\caption{Instance segmentation performance using real, synthetic and augmented datasets of various sizes tested on KITTI-15. (a) We fix the number of augmentations per image to 20 but vary the number of real image used for augmentation. This leads to a various size dataset depending on the number of real images.
	(b) We vary the number real images while keeping the resulting augmented dataset size fixed to 4000 images by changing the number of augmentations accordingly.
	(c) We train on various number of real images only. 
	(d) We train on various number of VKITTI images. }
	\label{fig:data_analysis} 
\end{figure*}

\begin{figure*}[t]
	\centering
	\begin{subfigure}[t]{0.245\textwidth}
		\centering
		\includegraphics[width=\textwidth]{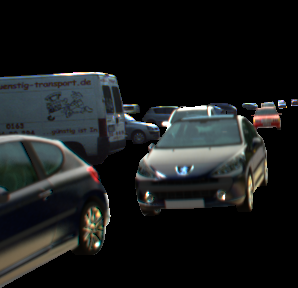}
		\caption{Black BG \\ AP50 = 21.5\%}
		\label{fig:black_bg}
	\end{subfigure}  
	\begin{subfigure}[t]{0.245\textwidth}
		\centering
		\includegraphics[width=\textwidth]{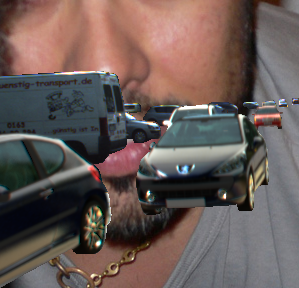}
		\caption{Flickr BG \\ AP50 = 40.3\%}
		\label{fig:flickr_bg}
	\end{subfigure}   
	\begin{subfigure}[t]{0.245\textwidth}
		\centering
		\includegraphics[width=\textwidth]{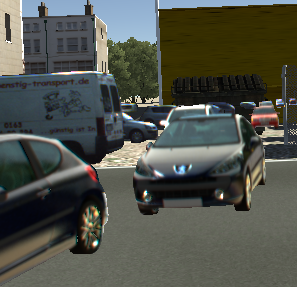}
		\caption{Virtual KITTI BG \\ AP50 = 47.7.3\%}
		\label{fig:vkitti_bg}
	\end{subfigure}  
	\begin{subfigure}[t]{0.245\textwidth}
		\includegraphics[width=\textwidth]{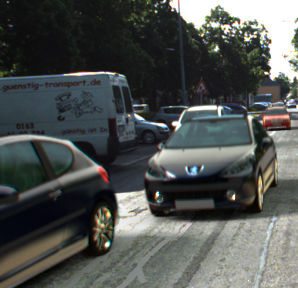}
		\caption{Real BG \\  AP50 = 49.7\%}
		\label{fig:real_bg}
	\end{subfigure}
	\caption{Comparison of performance of models trained on augmented foreground  cars (real and synthetic) over different kinds of background.}
	\label{fig:bg_comp}
\end{figure*}

\begin{figure*}[t]
	\centering
	\begin{subfigure}[t]{0.245\textwidth}
		\includegraphics[width=\textwidth]{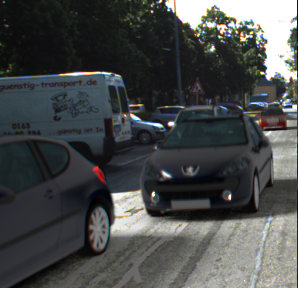}
		\caption{No env. map \\  AP50 = 49.1\%}
		\label{fig:no_env}
	\end{subfigure}
	\begin{subfigure}[t]{0.245\textwidth}
		\centering
		\includegraphics[width=\textwidth]{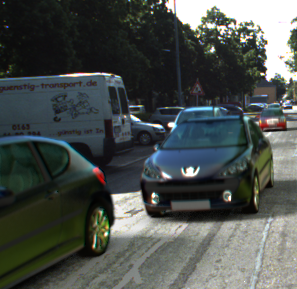}
		\caption{Random env. map \\ AP50 = 49.2\%}
		\label{fig:rand_env}
	\end{subfigure}
	\begin{subfigure}[t]{0.245\textwidth}
		\centering
		\includegraphics[width=\textwidth]{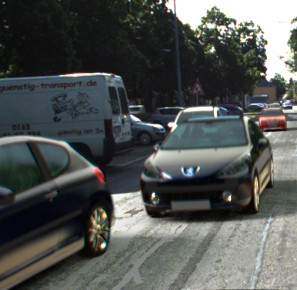}
		\caption{True env. map \\ AP50 = 49.7\%}
		\label{fig:true_env}
	\end{subfigure} 
	\begin{subfigure}[t]{0.245\textwidth}
		\centering
		\includegraphics[width=\textwidth]{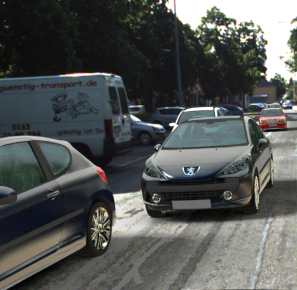}
		\caption{No postprocessing\\AP50 = 43.8\%}
		\label{fig:raw_nopost}
	\end{subfigure} 
	\caption{Comparison of the effect of post-processing and environment maps for rendering.}
	\label{fig:env_comp}
\end{figure*}

\begin{figure}[t]
	\centering
	\includegraphics[width=0.5\textwidth]{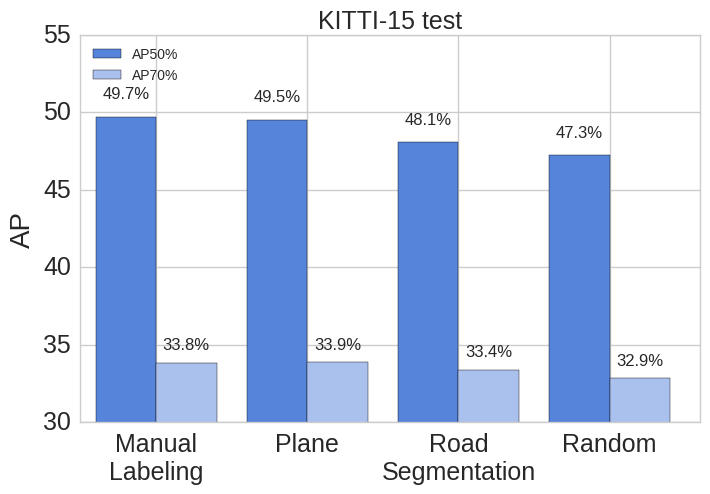}
	\caption{Results using different techniques for sampling car poses.}
	\label{fig:car_location}
\end{figure}

The potential usefulness of data augmentations comes mainly from its ability to realistically expand a relatively small dataset and train more generalizable models. We analyze here the impact of dataset size on training using real, synthetic and augmented data. 
Figures \ref{fig:num_real_aug} and \ref{fig:num_real} show the results obtained by training on various number of real images with and without augmentation, respectively. 
The models trained on a small real dataset suffer from over-fitting that leads to low performance, but then slowly improve when adding more training images. Meanwhile, the augmented datasets reach good performance even with a small number of real images and significantly improve when increasing dataset size outperforming the full real data by a large margin.
This suggests that our data augmentation can help improve the performance of not only smaller datasets, but also medium or even larger ones. 
\\
In figure \ref{fig:num_bg}, the total size of the augmented dataset is fixed to 4000 images by adjusting the number of augmentations for each real dataset size. In this case the number of synthetic car instances is equal across all variants which only differ in the number of real backgrounds.
The results highlight the crucial role of the real background diversity in the quality of the trained models regardless of the number of added synthetic cars.
\\
Even though fully synthetic data generation methods can theoretically render an unlimited number of training images, the performance gain becomes smaller as the dataset grows larger.   
We see this effect in figure \ref{fig:num_vkitti} where we train the model using various randomly selected subsets of the original VKITTI dataset. In this case, rendering adding data beyond 4000 images doesn't improve the model performance.

\subsection{Realism and Rendering Quality}

Even though our task is mainly concerned with segmenting foreground car instances, having a realistic background is very important for learning good models. Here, we analyze the effect of realism of the background for our task. In Figure \ref{fig:bg_comp} we compare models trained on the same foreground objects consisting of a mix of real and synthetic cars, while changing the background using the following four variations: (i) black background, (ii) random Flickr images (\cite{Philbin2007CVPR}), (iii) Virtual KITTI images, (iv) real background images. 
The results clearly show the important role of the background imagery and its impact even when using the same foreground instance. Having the same black background in all training images leads to over-fitting to the background and consequently poor performance on the real test data. Using random Flickr images improves the performance by preventing background over-fitting but fails to provide any meaningful semantic cues for the model. VKITTI images provide better context for foreground cars improving the segmentation. Nevertheless, it falls short on performance because of the appearance difference between the foreground and background compared to using real backgrounds. 

Finally, we take a closer look at the importance of realism in the augmented data. In particular, we focus on three key aspects of realism that is, accurate reflections, post-processing and object positioning. Reflections are extremely important for visual quality when rendering photo-realistic car models (see Figure \ref{fig:env_comp}) but are they of the same importance for learning instance-level segmentation? In Figure \ref{fig:env_comp} we compare augmented data using the true environment map to that using a random environment map chosen from the same car driving sequence or using no environment map at all. The results demonstrate that the choice of environment map during data augmentation affects the performance of the instance segmentation model only minimally.
This finding means that it's possible to use our data augmentation method even on datasets that do not provide spherical views for the creation of accurate environment map.
On the other hand, comparing the results with and without post-processing (Figure \ref{fig:true_env}+\ref{fig:raw_nopost}) reveals the importance of realism in low-level appearance.
 \\
Another important aspect which can bias the distribution of the augmented dataset is the placement of the synthetic cars. We experiment with 4 variants: (i) randomly placing the cars in the 3D scene with random 3D rotation, (ii) randomly placing the cars on the ground plane with a random rotation around the up axis, (iii) using semantic segmentation to find road pixels and projecting them onto the 3D ground plane while setting the rotation around the up axis at random, (iv) using manually annotated tracks from birdseye views. Figure \ref{fig:car_location} shows our results. Randomly placing the cars in 3D performs noticeably worse than placing them on the ground plane. This is not surprising as cars  can be placed at physically implausible locations, which do not appear in our validation data. The road segmentation method tends to place more synthetic cars in the clear road areas closer to the camera which covers the majority of the smaller (real) cars in the background leading to slightly worse results. The other 2 location sampling protocols don't show significant differences. This indicates that manual annotations are not necessary for placing the augmented cars as long as the ground plane and camera parameters are known.

\section{Conclusion} 
\label{sec:conclusion}
In this paper, we have proposed a new paradigm for efficiently enlarging existing data distributions using augmented reality. The realism of our augmented images rivals the realism of the input data, thereby enabling us to create highly realistic data sets which are suitable for training deep neural networks. In the future we plan to expand our method to other data sets and training tasks. We also plan to improve the realism of our method by making use of additional labels such as depth and optical flow or by training a generative adversarial method which allows for further fine-tuning the low-level image statistics to the distribution of real-world imagery.

\bibliographystyle{ieee}
\bibliography{bibliography_long,bibliography,bibliography_manual}

\end{document}